\documentclass[leqno,, 12pt]{article}
\usepackage{times}
\usepackage{filecontents}
\usepackage{amsmath,graphicx,graphics,epsfig,amsmath,amsfonts,amssymb,color}
\usepackage{latexsym}
\usepackage{setspace}
\usepackage{parskip}

\usepackage{graphics,amsmath,amsfonts,amssymb,epsfig}
\usepackage{latexsym}
\usepackage{amsfonts}
\usepackage{amsmath}
\usepackage{graphicx}
\usepackage{amssymb}
\usepackage{color}
\usepackage[]{algorithm2e}
\usepackage{cite}

\usepackage{deluxetable}
\usepackage{caption}
\usepackage{authblk}

\usepackage[margin= 1 in]{geometry}
\usepackage{setspace}
\usepackage{fancyhdr}
\newcommand{\bftheta}{{\mbox{\boldmath $\theta$}}}
\newcommand{\bfbeta}{{\mbox{\boldmath $\beta$}}}

\newcommand{\bfSigma}{{\mbox{\boldmath $\Sigma$}}}
\newcommand{\bfalpha}{{\mbox{\boldmath $\alpha$}}}

\newcommand{\bfmu}{{\mbox{\boldmath $\mu$}}}

\newcommand{\bfx}{\mathbf{x}}
\newcommand{\bfy}{\mathbf{y}}
\newcommand{\bfz}{\mathbf{z}}
\newcommand{\bfw}{\mathbf{w}}

\newcommand{\bfZ}{\mathbf{Z}}
\newcommand{\bfW}{\mathbf{W}}

\newcommand{\bfeta}{ \boldsymbol{\eta} }
\newcommand{\bflambda}{ \boldsymbol{\lambda} }
\newcommand{\bfgamma}{ \boldsymbol{\gamma} }

\newcommand{\beq}{\begin{equation}}
\newcommand{\eeq}{\end{equation}}

\newcommand{\beqs}{\begin{equation*}}
\newcommand{\eeqs}{\end{equation*}}

\newcommand{\beqa}{\begin{eqnarray}}
\newcommand{\eeqa}{\end{eqnarray}}

\newcommand{\bpm}{\begin{pmatrix}}
\newcommand{\epm}{\end{pmatrix}}

\newcommand{\argmin}{\operatornamewithlimits{argmin}}
\newcommand{\argmax}{\operatornamewithlimits{argmax}}

\title{Rare Disease Physician Targeting: A Factor Graph Approach}
\date{}

\author{Yong Cai}
\author{Yunlong Wang\thanks{yunlong.Wang@us.imshealth.com}}
\author{Dong Dai}
\affil{Advanced Analytics Department, QuintilesIMS}

\begin{document}

\maketitle

\begin{abstract}
In rare disease physician targeting, a major challenge is how to identify physicians who are treating diagnosed or underdiagnosed rare diseases patients. Rare diseases have extremely low incidence rate. For a specified rare disease, only a small number of patients are affected and a fractional of physicians are involved. The existing targeting methodologies, such as segmentation and profiling, are developed under mass market assumption. They are not suitable for rare disease market where the target classes are extremely imbalanced. The authors propose a graphical model approach to predict targets by jointly modeling physician and patient features from different data spaces and utilizing the extra relational information. Through an empirical example with medical claim and prescription data, the proposed approach demonstrates better accuracy in finding target physicians. The graph representation also provides visual interpretability of relationship among physicians and patients. The model can be extended to incorporate more complex dependency structures. This article contributes to the literature of exploring the benefit of utilizing relational dependencies among entities in healthcare industry.


\end{abstract}

\section*{Introduction}
A rare disease, also known as orphan disease, has very low prevalence rate that affects only a small percentage of population. In some extreme cases, for an example, Hutchinson-Gilford progeria syndrome only affects a few dozen children \cite{field2011rare}. Given the low patient number and high cost of bringing new product into the market, it is essential for the pharmaceutical companies to develop budget friendly and efficient marketing approaches.

Pharmaceutical companies use communication channels such as in-personal detailing or non-personal digital channels to raise disease awareness and deliver promotional messages to physicians \cite{narayanan2009heterogeneous,dong2009quantifying,manchanda2004response}. Detailing channel is also used for patient education and health promotion. There is literature to show that academic detailing can help increase disease detection and provide early disease intervention \cite{cameron2010evaluation, fox2008improving}.  Delivering educational messages related to the rare disease diagnosis and treatment can raise disease awareness and help diagnosis. When to launch a new orphan drug into the market, one practical question arises: under limited budget and resource, how can managers target only those relevant physicians instead of reaching out to vast majority? Many physicians treating rare disease cannot be directly identifiable from database containing diagnosis or prescription treatment data. Marketers need to use some predictive analytics to identify those individuals. On the other hand, it’s such a small affected population compared to that in common condition diseases. Finding undiagnosed patients and potential physicians is like looking for a needle in a haystack.

To identify physicians having rare disease patients is a challenging task from many perspectives. First of all, some of the rare diseases are very hard to diagnose. Patients affected by rare diseases may be free of symptoms for a long time or the symptoms can be hidden behind common conditions \cite{de2013update}. On top of that, physicians rarely encounter such patients in their daily practice and have limited experience or knowledge. They may be unaware that they have undiagnosed patients with such rare diseases. Even for the diagnosed patients, the medical record database may not capture all information because of the coverage or missing diagnosis codes in the system. These existing targets cannot be directly identified. But every potential target is valuable because of the market itself is very small.

The current approaches for identifying targets in database marketing or target marketing are developed under assumption for large markets \cite{hughes2005strategic, van2003predicting}. They prioritize customers by defined value. For example, one method is to derive customer targets through segmentation or clustering framework \cite{desarbo2008clusterwise}. These approaches don’t perform well in rare disease market where the class of interest is small and extremely imbalanced \cite{akbani2004applying}. The physicians, especially primary care physicians, who treat rare disease patients have similar characteristics or patient profiles as other physicians. Segmentation and profiling methods group all look-alike physicians together and do not differentiate well for the true rare disease physicians. The researchers can also use supervised classification approaches. But the traditional classification models have difficulties to predict smaller classes well \cite{chawla2009data}.

The objective of this article is to explore new ways to improve the targeting accuracy in the unique rare disease markets. Our motivation comes from the desire to enhance unsatisfactory results in the rare disease targeting practice. The traditional statistical models yielded targeting list with high false positive rate. One needed to reach out to a larger number of non-targeted physicians in order to cover some true rare disease physicians. Those models cannot effectively utilize dependencies among entities. We hope to improve the targeting accuracy in a new model by explicitly using the extra physician and patient relationship. In our rare disease targeting demonstration, there are two distinct feature spaces: patient and physician. These two spaces can be bridged by physician-patient treating relationship. But features and data dimensionalities from these spaces are completely different. A physician typically treats multiple patients with various conditions. The traditional classification model requires aggregating patient data into physician level before building physician classification model. Similarly, one can also build patient level prediction model and then link patient predicted flags to physician targets in a separate step. But either of the methods throw away relational information. Due to the extremely imbalanced classes, these models tend to generate high false positive predictions.

We propose a graphical model method to structurally model physician patient features together and utilize the additional relational information to improve target identification accuracy. Our hope is that the information from dependencies among physicians, patients, and between physician-patient can contribute to the accuracy gain. We first formulate the physician classification problem in a probabilistic joint distribution. The proposed model depicts the dependence structure among physicians and patients and relaxes i.i.d. assumptions. Then we use factor graph message passing algorithm to predict physician and patient labels.

We organized the remaining of the article in the following way: in the next section, we discuss background of rare disease and review some related work. Then we describe the data source used for model development. Next we formulate the problem using graph representation that designed for rare disease physician identification. With the proposed model, we present the factor graph algorithm for target label prediction, and parameter estimation. Finally, experiments with real data as well as concluding remarks will be given in the last two sections.

\section*{Related Work}

The United States Rare Disease Act of 2002 defines it as a disorder affects fewer than 200,000 people. Other countries use similar definitions for example, European Organisation for Rare Diseases \cite{EURORDIS2005} defines rare disease prevalence rate to be less than 1 in 2,000 people. The National Organization for Rare Disorders (NORD) at the National Institutes of Health (NIH) identified about 7,000 rare diseases. Collectively, rare diseases can affect 25-30 million Americans \cite{nord2013}. Developing orphan drugs for rare diseases represents a unique opportunity to pharmaceutical industry. In marketing, rare diseases bring different challenges than mainstream products. The healthcare environment calls for innovative methodologies to address these challenges.

Many rare disease patients are undiagnosed or misdiagnosed. To identify these patients and their treating physicians using predictive model is a challenging task. One major challenge comes from the imbalance of the classes in the dataset. Classic statistical models or standard machine learning algorithms are biased toward larger classes in prediction. If not treating and measuring properly, most of rare disease patients will be mis-classified as the other major classes albeit the overall accuracy rate may appear to be high. Oversampling and undersampling are commonly used techniques to overcome the imbalance problems \cite{chawla2009data, rahman2013addressing}. The focus of these algorithms is to boost signal and reduce prediction bias by reusing existing samples. Our proposed models address the problem from different angle. Instead of modeling separately in patient or physician space, we develop probabilistic graphical models to take advantage of the extra relational information among different entities.

There are very few literatures addressing rare disease physician targeting challenge. Some use predictive classification model to identify targets. \cite{chawla2013bringing} proposed using collaborative filtering to predict personalized disease based on patient history, phenotype and comorbidities. Collaborative filtering leverages the similarity among patients to profile disease risk for each individual patient. \cite{santoro2015rare} used hierarchical clustering to characterize and identify rare disease topologies. They apply random forests to derive the most important variables for profiling. The existing rare disease classification method works in either patient or physician data space. We propose a graphical model method to explicitly link physician and patient features.

Probabilistic graphical model \cite{bishop2006pattern, koller2009probabilistic} uses graphical diagram to visualize the statistical dependence among random variables. It encodes the problem into a joint probabilistic distribution over a high dimensional space. In a complex system, the statistical inference is computational demanding. Because it requires multidimensional integration for unknown variables. Factor graph \cite{kschischang2001factor}, a major class in graphical models, can explain the dependencies among interacting variables. By factoring the global multivariate functions into several local functions, the factor graph can efficiently perform statistical inference through messaging passing algorithm. It is widely applied in statistical learning, signal processing and artificial intelligence \cite{loeliger2007factor}. In the next sections, we will describe the data assets, formulate the probabilistic model and develop factor graph for  physician targets prediction.

\section*{Data Description}
We extract data from  IMS Health longitudinal prescription (Rx) and medical claims (Dx) database. To limit the scope, we focus on only one particular rare disease market. The selected rare disease is an inherited blood disorder caused by genetic defect. It is estimated to affect about 1 in 50,000 people according to Genetic Home Reference (GHR) from NIH.

The Rx data is derived from electronic records collected from pharmacies, payers, software providers and transactional clearinghouses.  This information represents activities that take place during the prescription transaction and contains information regarding the product, provider, payer and geography. The Rx data is longitudinally linked back to an anonymous patient token and can be linked to events within the data set itself and across other patient data assets. Common attributes and metrics within the Rx data include payer, payer types, product information, age, gender, 3-digit zip as well as the scripts relevant information including date of service, refill number, quantity dispensed and day supply.  Additionally, prescription information can be linked to office based claims data to obtain patient diagnosis information. The Rx data covers up to 88\% for the retail channel, 48\% for traditional mail order, and 40\% for specialty mail order.

The Dx data is electronic medical claims from office-based individual professionals, ambulatory, and general health care sites per year including patient level diagnosis and procedure information.  The information represents nearly 65\% of all electronically filed medical claims in the US.  All data is anonymous at the patient level and HIPAA compliant to protect patient privacy.

For model development, we pull the diagnoses, procedures and prescriptions at transaction level using study period from January 1, 2010 to July 31, 2015. In the rest of the article, we name a patient with the rare disease condition as ``positive patient" and name the rest of them as ``negative patient". Similarly, let ``positive physician" be a physician who treats at least one positive patient, and ``negative physician" be a physician treating only negative patients.

From the extracted data, we can positively identify 1,233 true rare disease patients with valid records such as gender, age and region. To boost the positive signal and model development, we construct a training and validation data by matching each positive patient with 200 randomly selected negative patients. The final patient data contains 1,233 positive patients and 246,600 negative patients. The positive ratio in the training data is about 0.5\%.

Based on the linkable anonymous IDs in the patient data, we further pull data of physicians who have treated those patients in predefined selection period. Physicians and patients are linked if they have associated in at least one medical claim record in the selection period.  We end up with 68,898 unique physicians in total and among those 8,346 positive physicians. There are 1,463,030 physician-patient links stored in a separate database. On average each patient has visits 5.9 physicians, and each physician has treated 21.23 patients.

Up to this point the experiment data exhibits some challenges. First, the imbalance among positive and negative classes limits the performance of many common machine learning and statistical models like regression, support vector machine and decision trees. Second, the complicated relationships between patients and physicians make it difficult to directly generate meaningful features as model input from raw data. Third, the large amount of data calls for an efficient inference algorithm instead of naive marginalization. We'll propose our model in the next sections.

\section*{Problem Formulation}

Given the data we just described, let us formally define the problem in mathematics. Consider a multi-agent system that consists of $N$ physicians $A_i$, and $M$ patients $B_j$, with $i\in {\mathcal N}_A=\{1,2,...,N\}$ and $j\in {\mathcal N}_B=\{1,2,...,M\}$ respectively. Each physician  $A_i$ is associated with a feature vector $\bfz_i \in \mathbb{R}^L$ representing  physician  $A_i$'s features such as her specialty, age, gender, office location, etc.  Similarly, each patient $B_j$ is associated with a feature vector $\bfw_j \in \mathbb{R}^K$ denoting her features such as age, gender, diagnosis histories, etc.  In a matrix form, let  matrix $\bfZ\in \mathbb{R}^{N \times L} = [\bfz_1, \bfz_2, \cdots, \bfz_N ]^\top$  summarizes all the features of all the physicians, and let matrix $\bfW \in \mathbb{R}^{N \times K} = [\bfw_1, \bfw_2, \cdots, \bfw_M ]^\top$  summarizes all the features of all the patients.

\begin{figure}[htb]
\centering
\includegraphics[width = 14cm]{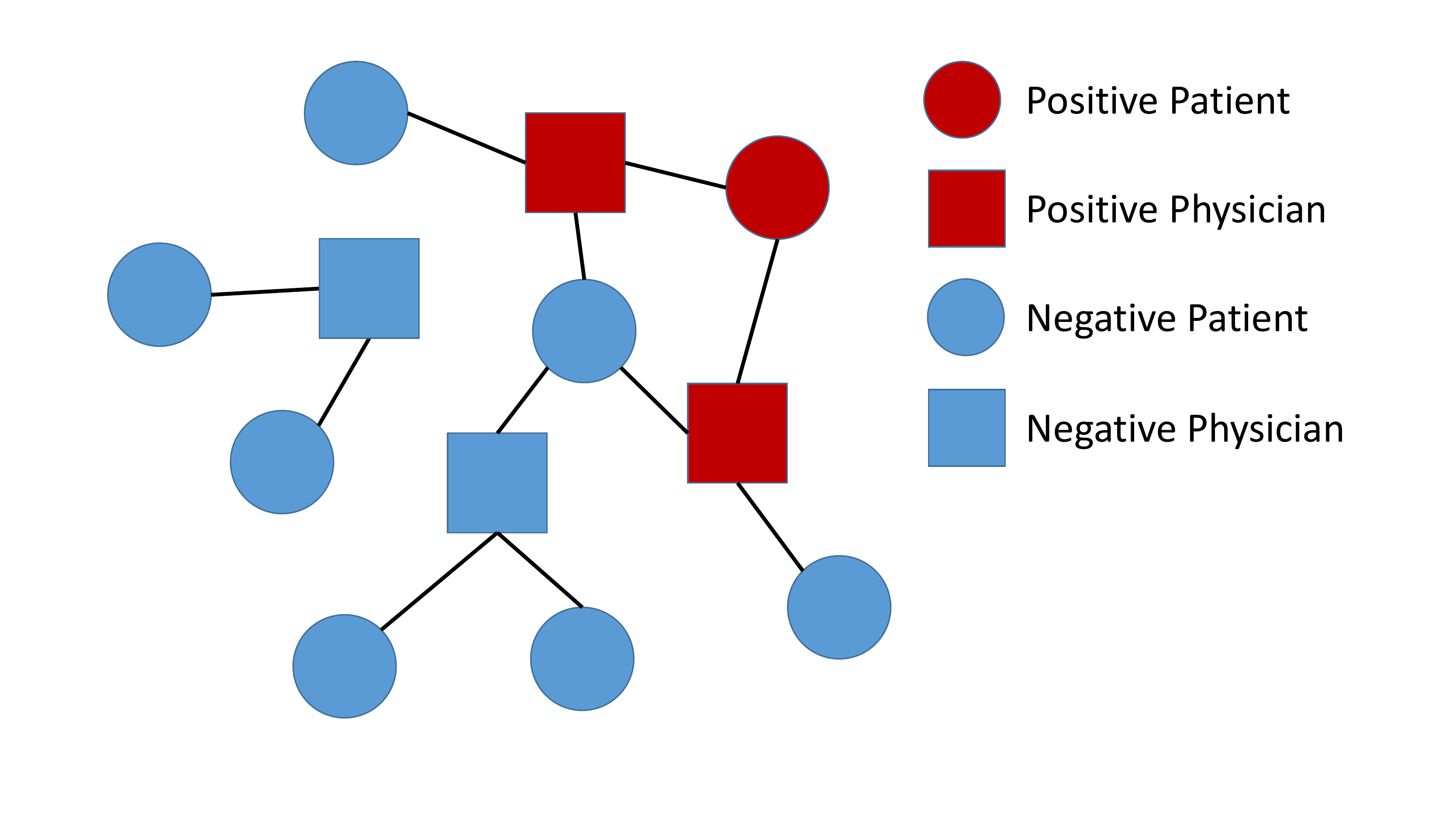}
\caption{ This schematic depicts the physician-patient network }
\label{schematic}
\end{figure}

Let $\bfx$ be patient label vector indicating if a patient has a specific rare disease. Specifically, $\bfx = [x_1, x_2,\cdots, x_M]^\top \in \{0, 1\}^M$ is a binary vector with 0 or 1 entries. If $x_j=1$ then patient $B_j$ is positive, vice versa. Similarly, let us define the physician label vector as $\bfy = [y_1, y_2, \cdots, y_N] \in \{0,1\}^N$.  Then by the definition of positive physician and negative physician, the studied physician-patient network can be illustrated as a graph $G = ([{\mathcal N}_A, {\mathcal N}_B], {\mathcal E})$ in Figure \ref{schematic}. In this graph, each square shape denotes a physician and each circle a patient; the red color denotes positive label and the blue negative label. Please note that other than the physician and patient's labels and features, the patient-to-physician relationship is also known from the data.

With these defined components, we can formulate the problem as follows. Given known patient features $\bfW$ , physician features $\bfZ$ and the physician-patient network $G$,  for all $i \in {\mathcal N}_A$, find the estimate of $y_i$ that minimizes the mean square error, referred to as the Minimum mean square error (MMSE) estimate. It can be shown that this MMSE estimate has the form
\begin{eqnarray}
\hat{y}_i &=& \argmin_{\tilde{y}_i} \mathbb{E} (\tilde{y}_i - y_i )^2\nonumber\\
&=& \mathbb{E}\left\{   y_i | \bfW , \, \bfZ   \right\}\nonumber
\end{eqnarray}
where the first  expectation  is taken over both  $\bfW$, $\bfZ$ and $y_i$, and the second expectation refers to the expected value of $y_i$ with posterior distribution $p(y_i | \bfW , \, \bfZ)$.

The goal is to find physician labels $\tilde{y}_i$ using patient and physician features such that the distance between $\tilde{y}_i$ and $y_i$ is minimized.  By building probabilistic models, one can compute the posterior distribution $p(y_i | \bfW , \, \bfZ)$ and the prediction of physician label is the MMSE of $y_i$.

\section*{Proposed Model with Feature Engineering}

Next we show the derivation of the proposed model. To account for the dependencies among physicians and patients, we use Bayes network \cite{friedman1997bayesian} to build a model representing the joint distribution of all observed and latent variables\footnote{Observed variables refer to the variables whose values are given, e.g., $\bfW$ and $\bfZ $. Oppositely, latent variables refer to the variables with unknown value, like $\bfx$ and $\bfy $ in the model.}. Specifically, we show the explicit formulas of the joint distribution, which serves as the basis of the following section where we will convert the Bayes network to a factor graph.

\begin{figure}[htb]
\centering
\includegraphics[width = 14cm]{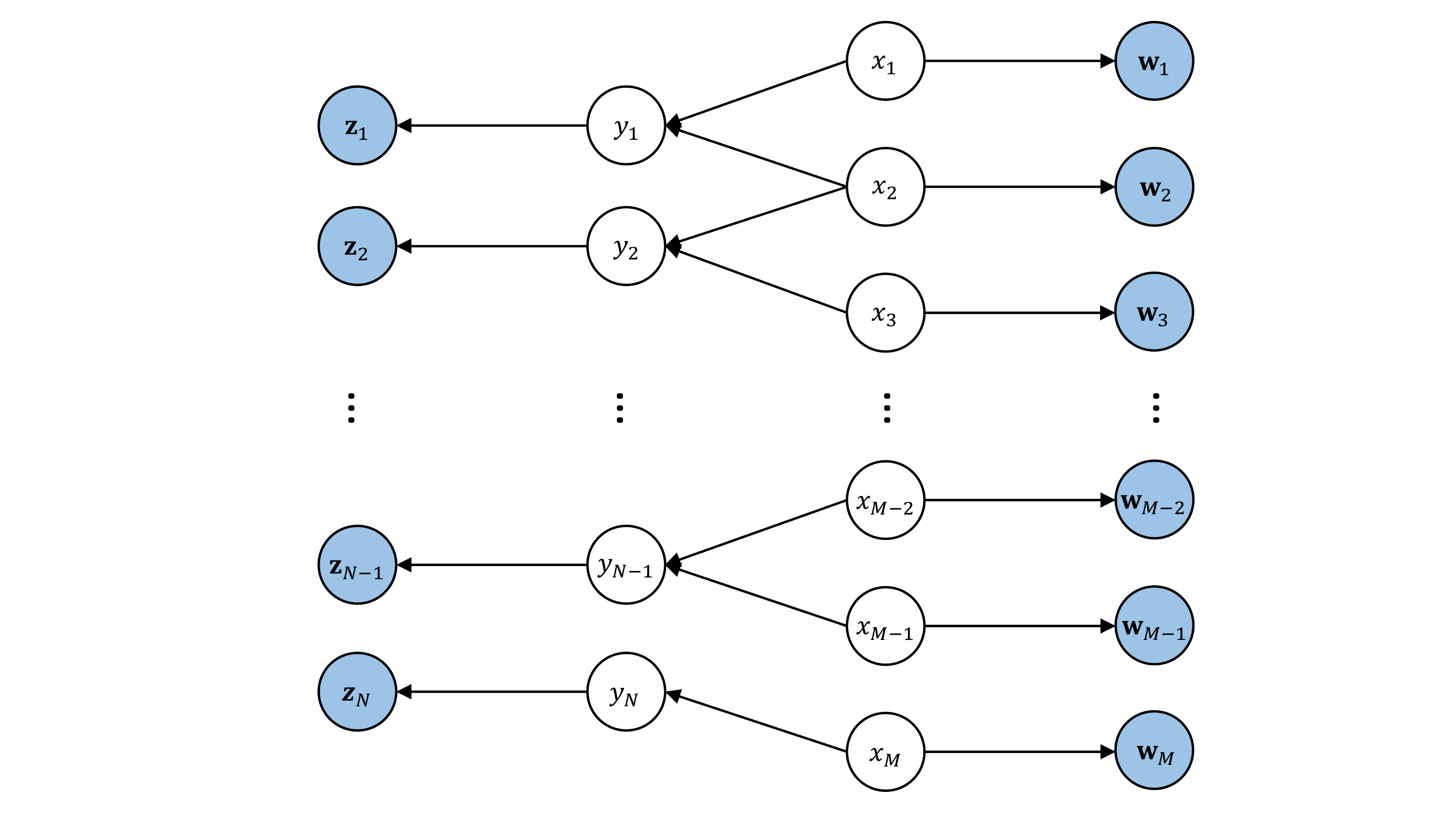}
\caption{ This graph depicts the relationship between random variables by employing Bayes network. Each node represents a random variable and each edges depicts a conditional probability distribution. $\bfy$ and $\bfZ$ represent physician labels and features; $\bfx$ and $\bfW$ represent patient labels and features, respectively. The whole graph yields a joint probability distribution of all the random variables.}
\label{JMR_BN}
\end{figure}

To depict the relationship between physicians and patients as well as the features and lables associated with them, we draw a four layer Bayes network in Figure \ref{JMR_BN}.  Using the definition from previous problem formulation, we can write the joint distribution of all the random variables in a factorized form as
\begin{eqnarray}\label{main}
p(\bfW, \bfx, \bfy, \bfZ) &=& p(\bfZ| \bfy) p(\bfy| \bfx)p(\bfW| \bfx)p( \bfx),
\end{eqnarray}
where physician features $\bfZ$ and patient features $\bfW$ are conditional independent given the corresponding physician or patient labels. Note that we remark the factors coming from either the patient side or the physician side.

In equation \ref{main}, $p(x)$ can be considered as a prior distribution of patient labels. Because the experiment data set is constructed by matching one positive patient with two hundred negative patients, we assign an a priori probability of one patient being positive to be $1/201$. Equivalently, we have
\begin{eqnarray}
p( \bfx) &=& \prod_{j = 1}^M  p( x_j) \nonumber\\
&=& \prod_{j = 1}^M \eta ^ {\mathbb{I}(x_j = 1)}  ( 1 - \eta )^ {\mathbb{I}(x_j = 0)},\nonumber
\end{eqnarray}
with $\eta = 1/201$ a priori.

In the following subsections, we will develop the formulas for the rest of the components  $p(\bfZ| \bfy)$, $p(\bfW| \bfx)$, and $p(\bfy| \bfx)$.

\subsubsection*{Patient Feature Distributions}

The patient feature distribution $p(\bfW| \bfx)$ summarizes the patient information, which includes patient labels, demographics, diagnoses, procedures and prescription treatments. The condition distribution $p(\bfW| \bfx)$ represents the patient feature distributions for both positive and negative classes. For clear demonstration, we divide the patient feature space into three categories:
\begin{itemize}
\item $\bfw_j^{S}$: patient $B_j$'s self features including $B_j$'s gender, age decade and region.
\item $\bfw_j^{L}\in \{0,1\}^{58}$: a fifty eight by one row vector representing patient $B_j$'s 58 clinical code indicators (Y/N). A clinical code can be either patient diagnosis, procedure or prescription filled.)\footnote{The size of the vector can be adjusted accordingly for other studies. 58 clinical codes here are specified for this experiment only.}.
\item $\bfw_j^{F}\in \mathbb{Z}^{58}$: a fifty eight by one row vector representing patient $B_j$'s clinical code frequency for the same 58 clinical codes. This feature represents how many times the event of prescription, procedure or diagnosis occurred during the study period.
\end{itemize}
By independence assumption between patients, we can get that
\begin{eqnarray}
p(\bfW| \bfx) &=& \prod_{j = 1}^M  p( \bfw_j |x_j) \nonumber\\
&=& \prod_{j = 1}^M  p( \bfw_j^{S} |x_i)p( \bfw_j^{L} |x_i)p( \bfw_j^{F} |\bfw_j^{L} , x_j)\nonumber
\end{eqnarray}

In order to get this factor form, we make two assumptions here. First, we assume that the features as well as labels are mutually independent among patients. Second, given a patient label $x_j$, this patient's self features $\bfw_j^{S}$ are independent with her clinical code features, i.e., $\bfw_j^{L}$ and $\bfw_j^{F}$.

Specifically, we model
\begin{eqnarray}\label{pat_self_feature}
p( \bfw_j^{S} |x_j) &=& \prod_{l = 1}^4  p( \bfw_{j,l}^{S} |x_j)
\end{eqnarray}
where $\forall j \in \mathcal{N}_B$,
\begin{itemize}
\item $\bfw_{j,1}^{S}\in \{0,1\}$: patient gender
\begin{eqnarray}
p( \bfw_{j,1}^{S} |x_j) \sim
\begin{cases} \mathrm{Ber}( \eta_g^{1} ) & \text{if } x_j=1, \nonumber\\
\mathrm{Ber}( \eta_g^{0} ) & \text {if } x_j=0. \nonumber
\end{cases}
\end{eqnarray}


\item $\bfw_{j,2}^{S} \in \{0,1,2,\cdots, 9\}$: patient age decade
\begin{eqnarray}
p( \bfw_{j,2}^{S} |x_j) \sim
\begin{cases} \mathrm{Cate}( \bfgamma_a^{1} ) & \text{if } x_j=1, \nonumber\\
\mathrm{Cate}( \bfgamma_a^{0} ) & \text {if } x_j=0.\nonumber
\end{cases}
\end{eqnarray}
where Cate($\bfgamma$) means categorical distribution parametrized by vector $\bfgamma$ and the element-wise sums of both $\bfgamma_a^{0}\in [0,1]^{10}$ and $\bfgamma_a^{1}\in [0,1]^{10}$ are one.

\item $\bfw_{j,3}^{S}\in \{1, 2, 3, 4\}$: patient region whose value corresponds to `SOUTH', `WEST', `MIDWEST', `NORTHEAST'
\begin{eqnarray}
p( \bfw_{j,3}^{S} |x_j) \sim
\begin{cases} \mathrm{Cate}( \bfgamma_e^{1} ) & \text{if } x_j=1, \nonumber\\
\mathrm{Cate}( \bfgamma_e^{0} ) & \text {if } x_j=0. \nonumber
\end{cases}
\end{eqnarray}
where $\bfgamma_a^{0} \in [0,1]^{4}$, and $\bfgamma_a^{1} \in [0,1]^{4}$ respectively.

\end{itemize}

For patient clinical code indicator and frequency features, $p( \bfw_j^{L} |x_i)$ and $p(\bfw_j^{F} |\bfw_j^{L} , x_i)$, we propose modeling by Bernoulli and Poisson distribution respectively. As the clinical codes have been grouped into 58 disjoint classes, here we further assume that
\begin{eqnarray}
p( \bfw_j^{L} |x_j) &=& \prod_{q=1} ^ {58} p( w_{j,q}^{L} |x_j),\nonumber\\
p(w_j^{F} |\bfw_j^{L} , x_j) &=& \prod_{q=1} ^ {58} p( w_{j,q}^{F} | w_{j,q}^{L}),\nonumber
\end{eqnarray}
where $w_{j,d}^{L}$ and $w_{j,d}^{F}$ denotes the $d$th element of vectors $\bfw_j^{L}$ and $\bfw_j^{F}$ respectively.

Let $\bfeta_d^0 = [\eta_{d,1}^0, \eta_{d,2}^0, \cdots, \eta_{d,N}^0]^\top \in [0,1]^{58}$ and  $\bfeta_d^1 = [\eta_{d,1}^1, \eta_{d,2}^1, \cdots, \eta_{d,N}^1]^\top \in [0,1]^{58}$ be the parameter vectors indicating the probability of getting positive clinical indicator for positive and negative patient classes respectively. Then we have the following distribution
\begin{eqnarray}
p( w_{j,q}^{L} |x_j) \sim
\begin{cases}
\mathrm{Ber}( \eta_{d,q}^1 ) & \text{if } x_j=1, \nonumber\\
\mathrm{Ber}( \eta_{d,q}^0 ) & \text{if } x_j=0.\nonumber
\end{cases}
\end{eqnarray}

To compute $p( w_{j,q}^{F} | w_{j,q}^{L})$, we propose $\forall j \in \mathcal{N}_B$,
\begin{eqnarray}
p( w_{j,q}^{F} | w_{j,q}^{L}) \sim
\begin{cases}
\mathrm{Poi}( \lambda_{d,q}^1 ) & \text{if } w_{j,q}^{L}=1, \nonumber\\
\mathrm{Poi}( \lambda_{d,q}^0 ) & \text{if } w_{j,q}^{L}=0. \nonumber
\end{cases}
\end{eqnarray}
where Poi($\lambda$) symbolizes a Poisson distribution parametrized by $\lambda$, with $\lambda_{d,q}^0 > 0$ and $\lambda_{d,q}^1 >0$.

\subsubsection*{Physician Feature Distributions}

Similar to patient feature formulation, we can create the conditional distributions of physician features given physician label $p(\bfZ| \bfy)$. We separate physician features $\bfZ$ into two parts. The first part, $\bfz^{S}$, accounts for physician general demographics such as specialty, gender, patient count, and the state where his or her office locates. The second part, $\bfz^{D}$, accounts for physician's overall office claims histories. For this part, we create maximum, minimum, sum and average of observed number of claims for each physician.

Specifically, we model
\begin{eqnarray}
p(\bfZ| \bfy) &=& \prod_{i = 1}^N  p( \bfz_i |y_i)\nonumber\\
&=& \prod_{i = 1}^N  p( \bfz_i^{S} |y_i)p( \bfz_i^{D} |y_i) \nonumber
\end{eqnarray}
where $\forall i \in \mathcal{N}_A$, the above distribution can further be factorized as
\begin{eqnarray}\label{fact}
 p( \bfz_i^{S} |y_i) &=& \prod_{l = 1}^3  p( \bfz_{i,l}^{S} |y_i).
\end{eqnarray}

In Equation \ref{fact}, we have

\begin{itemize}
\item physician gender $\bfz_{i,1}^{S}\in \{0,1\}$:  $p( \bfz_{i,1}^{S} |y_i) \sim \mathrm{Ber}( \omega_g )$, where if $y_i = 1$, $\omega_g = \omega_g^{(1)}$, else $\omega_g = \omega_g^{(0)}$.

\item physician specialty code $\bfz_{i,2}^{S}\in \{0,1,\cdots, 189 \}$:  $p( \bfz_{i,2}^{S} |y_i) \sim \mathrm{cate}( \bfgamma_p )$

\item physician patient count $\bfz_{i,3}^{S}\in \mathbb{N}^+ $:  $p( \bfz_{i,3}^{S} |y_i) \sim \mathrm{Pois}( \lambda_c )$

\end{itemize}
where in all of above distributions, we will have separate sets of parameters for both positive and negative classes.

Let $z_{i,l}^{D} \in \mathbb{R} $ be the maximum, minimum, summation, and average number of office claims of physician $i$ for his/her patients, for $l \in \{1,2,3,4\}$. Also let $\bfz^{D} = [z_{i,1}^{D},z_{i,2}^{D},z_{i,3}^{D},z_{i,4}^{D} ]^\top$ symbolize all the claims related features, then we assume that the claims related features follow a joint Gaussian distribution represented by,
\begin{eqnarray}
p( \bfz^{D} |y_i) \sim
\begin{cases}
{\cal N}( \bfmu_d^1, \bfSigma_D^1 ) & \text{if } y_i=1, \nonumber\\
{\cal N}( \bfmu_d^0, \bfSigma_D^0 ) & \text{if } y_i=0. \nonumber
\end{cases}
\end{eqnarray}
where ${\cal N}(\bfmu, \bfSigma)$ denotes multivariate normal distribution parametrized by $\bfmu$ and $\bfSigma$.

\subsubsection*{Physician Labels Joint Posterior Distribution}

The last component in the joint distribution is $p(\bfy|\bfx)$, which represents the conditional distribution of physician labels given patient labels. It describes the relationships between patient labels and physician labels from a probabilistic point of view. Intuitively, $p(\bfy|\bfx)$ demonstrates under what conditions a physician can be regarded as positive or negative. Let $\bfx^i$ be the positive labels of the patients associated with physician $A_i$. Note that physician labels are determined by her patient labels only, then the $p(\bfy|\bfx)$ can be factorized as
\begin{eqnarray}\label{eq:M.3.1}
p(\bfy|\bfx) &=& \prod_{i=1}^N p(y_i|\bfx) \nonumber \\
&=& \prod_{i=1}^N p(y_i|\bfx^i)
\end{eqnarray}
where the second equal sign is due to the fact that one physician's label is conditional independent with the labels of patients not in her set given her patients' labels, i.e., $p(y_i, \bfx^j|\bfx^i) = p(y_i|\bfx^i)p(\bfx^j|\bfx^i)$, $\forall j \neq i$.

In Equation \ref{eq:M.3.1}, the $p(y_i|\bfx^i)$ is given by
\begin{eqnarray}
p(y_i|\bfx^i)  &=& \left( 1 - \prod_{ j \in \mathcal{N}_i }\mathbb{I}(x_j = 0) \right)^ {1 - y_i}\left(  \prod_{ j \in \mathcal{N}_i }\mathbb{I}(x_j = 0) \right)^ {y_i}
\end{eqnarray}

\section*{Predictive Inference for Physician Targets}

So far we have derived explicit formulas for each factor form in Equation \ref{main}. The joint distribution  $p(\bfx, \bfy, \bfW, \bfZ)$ is just the product of all factors. In this section, we will follow the joint distribution formula to predict physician labels, $\bfy$. To compute the predictive results,
we first convert the Bayes network to a factor graph and then apply the prediction algorithm of inferring the physician labels given known features. We'll show the rationale and details below.

According to the Bayes rule, the joint posterior of $\bfy$ is given by
\begin{eqnarray}\label{P1}
p(\bfy|\bfW, \bfZ)  &=& \dfrac{1}{Z}  p(\bfy, \bfW, \bfZ)\\
 &=& \dfrac{1}{Z} \sum_{x_1 = 0}^1 \sum_{x_2 = 0}^1\cdots \sum_{x_M = 0}^1 p(\bfx, \bfy, \bfW, \bfZ), \nonumber
\end{eqnarray}
where $Z = \int_{\bfW}\int_{\bfZ} p(\bfx, \bfy, \bfW, \bfZ) d\bfZ d\bfW$ denotes the normalizing constant for posterior of $\bfy$. Here we remark that the $Z$ is computationally expensive but it will be shown soon that this constant doest not need to be computed. The $M$ order summation after the second equal sign is because one need to marginalize  $\bfx$ to compute $p(\bfy|\bfW, \bfZ)$.  Following Equation \ref{P1}, the marginalized posterior of $y_i$ has the form
\begin{eqnarray}\label{P2}
p(y_i|\bfW, \bfZ)  &=&  \sum_{y_{-i} = 0}^1  p(\bfy|\bfW, \bfZ),
\end{eqnarray}
where $\sum_{y_{-i} = 0}^1$ denotes a summation over all random variables expect $y_i$ in $\bfy$.

Note that computing $p(y_i|\bfW, \bfZ)$ implemented in Equation \ref{P1} and Equation \ref{P2} with brutal force will require marginalizing $M+N-1$ binary variables, whose computation cost increases exponentially as $N$ or $M$ grows. In our studied data set, we have $N = 68,898$ and $M = 247,833$, which prohibits us from using traditional marginalization methods. To compute the $p(y_i|\bfW, \bfZ)$ in an efficient manner, we introduce the factor graph model and its associated variable marginalizing algorithm which is called message passing algorithm.

\subsubsection*{Factor Graph and Message-passing algorithm}

\begin{figure}[htb]
 \centerline
 {\includegraphics[width = 14 cm]{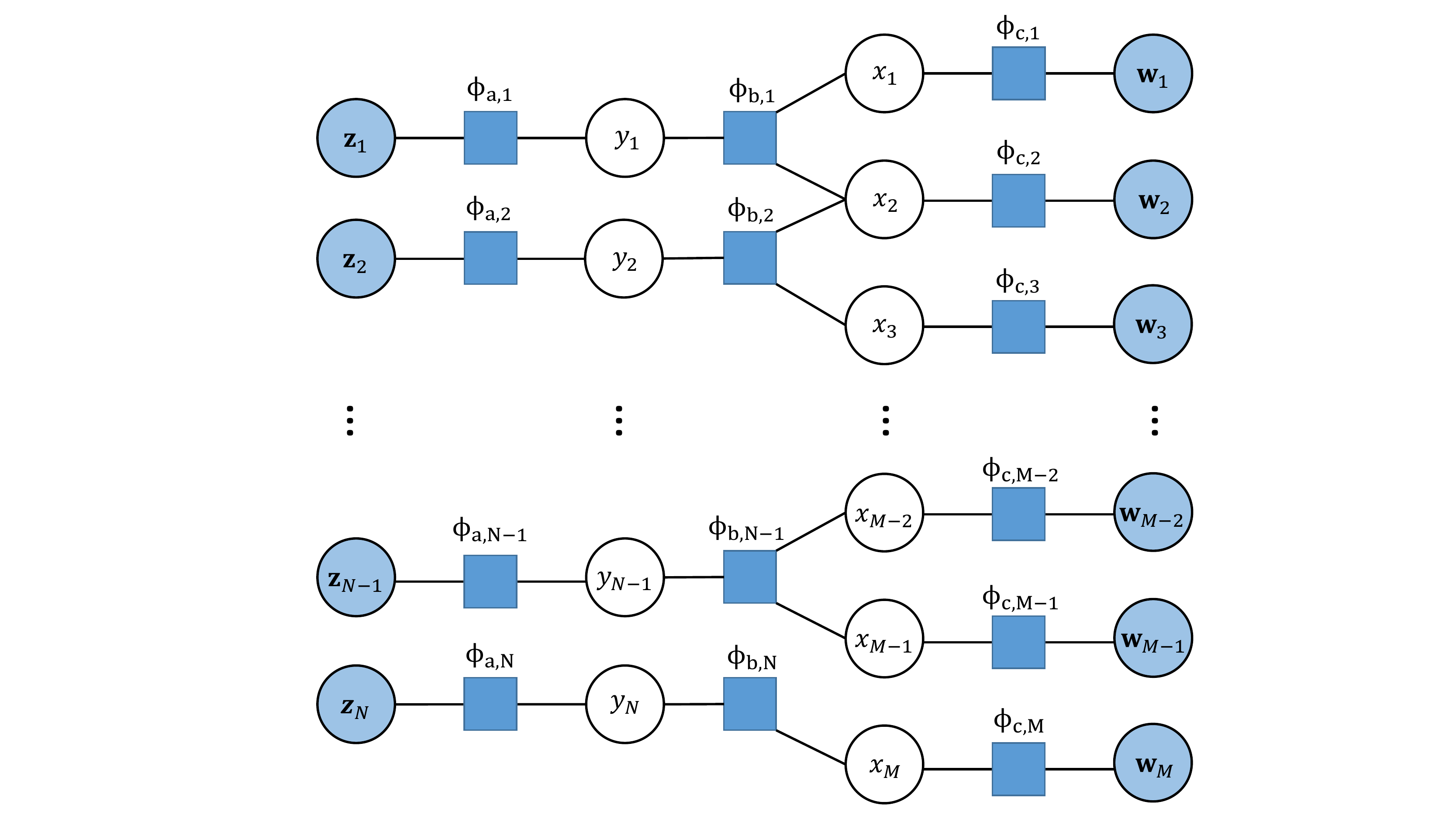}}
 \caption{ Factor graph depicting the relationships between variables nodes and factor nodes. }
 \label{JMR_FG}
\end{figure}

In our proposed model, the doctor labels are conditional independent with each other given the patient labels; and the patient labels are mutually conditional independent if the doctor labels are given. Due to such conditional independence in the proposed model, a factor graph will allow us to solve the original giant marginalization problem by exploit the ``Divide and Conqure'' strategy, which will be shown in the rest of the sections.

A factor graph \cite{kschischang2001factor} is a type of probabilistic graphical model that contains two types of nodes: Variables, which can be either evidence variables whose value is known, or query variables whose value should be predicted or marginalized. Factors, which define the relationships between variables in the graph. In the proposed model, patient features $\bfW$ and physician features $\bfZ$ are evidence variables while patient labels $\bfx$ and physician labels $\bfy$ are query variables.

The factor graph of the proposed model is drawn in Figure \ref{JMR_FG}, where each circle represents a variable node and each square represents a factor node. For the variable nodes, they have the exact same definition as the variables in the Bayes network in Figure \ref{JMR_BN}. For the factor nodes, there are three classes of them, $\phi_a$, $\phi_b$ and $\phi_c$. For each $i \in {\cal N}_A$, $\phi_{a,i}$ connects $\bfz_i$ and $y_i$, $\phi_{b,i}$ connects $y_i$ and the $x_j$, where $j\in {\cal N}_i$. For each $j \in {\cal N}_B$, $\phi_{c,j}$ connects $x_j$ and the $\bfW_j$.

By noting the factorization form in Equation \ref{main}, we can define the specific definitions of $\phi_a$, $\phi_b$ and $\phi_c$ by
\begin{eqnarray}
\phi_{a,i}(\bfz_i, y_i) &=&  p(\bfz_i | y_i), \,\,\,\,\,\, \forall i \in {\cal N}_A,    \nonumber\\
\phi_{b,i}(y_i, \bfx^i) &=&  p(y_i, \bfx^i), \,\,\,\, \forall i \in {\cal N}_A,    \nonumber\\
\phi_{c,j}(\bfw_j, x_j) &=&  p(\bfw_j | x_j), \,\,\, \forall j \in {\cal N}_B.   \nonumber
\end{eqnarray}
Using the explicit forms of $p(\bfz_i | y_i)$, $p(y_i, \bfx^i)$ and $p(\bfw_j | x_j)$, one can show that the joint distribution $p(\bfx, \bfy, \bfW, \bfZ)$ can be factorized as:
\begin{eqnarray}\label{factor}
p(\bfx, \bfy, \bfW, \bfZ) = \prod_{i = 1}^N \phi_{a,i}(\bfz_i, y_i)\phi_{b,i}(y_i, \bfx^i) \prod_{j = 1}^M \phi_{c,j}(\bfw_j, x_j)
\end{eqnarray}

With the factor graph for our proposed model in Figure \ref{JMR_FG}, now we can apply the Message-passing algorithm \cite{kschischang2001factor} for marginalization. The topology of the factor graph in Figure \ref{JMR_FG} has no loop and it is a tree structure graph. Considering that \cite{kschischang2001factor} has shown exact marginalization results will be achieved in graphs without loop, we can get the exact value of $p(y_i|\bfW, \bfZ$) by the Message-passing algorithm.

Message-passing algorithms that operate on factor graphs aiming at calculating the marginal distribution for each unobserved node, conditional on any observed nodes. Here, we will show how to compute $p(y_i, \bfW, \bfZ)$ by the sum-product message-passing algorithm. The idea is that instead of computing $p(y_i|\bfW, \bfZ)$ by Equations \ref{P1} and \ref{P2} directly, it is much more efficient to first solve the $p(y_i, \bfW, \bfZ)$ by marginalizing all $\bfx$ and $\bfy_{-i}$ variables using sum-product algorithm, then compute
\begin{eqnarray}
p(y_i|\bfW, \bfZ) &=& \dfrac{p(y_i,\bfW, \bfZ)}{\sum_{y_i = 0}^1 p(y_i,\bfW, \bfZ)}  \nonumber
\end{eqnarray}
where the summation merely takes over a binary variable $y_i$.

As is shown by the factor graph topology in Figure \ref{JMR_FG}, it has more than one components, whereas each components of it has a tree structure. \cite{kschischang2001factor} show that the sum-product algorithm yields an exact result of $p(y_i, \bfW, \bfZ)$ by exact $2L$ message passings, where $L$ is the number of edges in the factor graph.  According to the algorithm, there are two types of messages. When the message is from a factor node $s$ to a variable node $v$, the message is a probability distribution given by
\begin{eqnarray}
\mu_{s\rightarrow v}(v) &=& \sum_{u\in {\cal N}_s \setminus v} \phi_s({\cal N}_s) \prod_{u\in {\cal N}_s \setminus v} \mu_{u\rightarrow s}(u),  \nonumber
\end{eqnarray}
where ${\cal N}_s$ represents all variable nodes in the neighbor set of node $s$, and node $u$ is a node in node $s$'s neighbor set but not equal to $v$. On the other hand, the message from a variable node $u$ to a factor node $s$ is given by
\begin{eqnarray}
\mu_{u\rightarrow s}(u) &=&  \prod_{\omega\in {\cal N}_u \setminus s} \mu_{\omega\rightarrow s}(u),  \nonumber
\end{eqnarray}
where $\omega$ denotes a node in node $u$'s neighbor set but not equal to $s$.

In the factor graph of the proposed method, one can show that the message from variable nodes to factor nodes are easy to compute because it is just the production of few functions. The computation cost of messages from factor nodes to variable nodes are seemingly large but one can show that when the summation with respect to variable $\bfw$ or $\bfz$, the summation result will be one. This is because $\phi_a$ and $\phi_c$ are probability density functions of  $\bfw$ and $\bfz$ respectively. When the summation is over variable $\bfx^i$ or $y_i$, the computation cost is still small because for every $\phi_b$ node, it only connects with a few variable nodes.

\section*{Parameter Learning}

Other than the predictive inference algorithm, we will show parameter estimation for the proposed model next. From machine learning point of view, the parameter estimation process can be understood as a training stage, during which the computer is trained to make meaningful predictions for the variables of interest, e.g.,the physician label in our problem. Following we will present the method and results of parameter estimation for the proposed model based on the training data set. In the remaining of this section, we assume that the values of the observed variables are from the training data set. In order to find the maximum likelihood estimate of the parameters, we aim at solving the following optimization problem:
\begin{eqnarray}\label{para_main}
\hat{\bftheta} &=& \argmax_{\bftheta} \, p(\bfW, \bfx,\bfy,  \bfZ;\bftheta ) \\
&=& \argmax_{\bfalpha, \bfbeta} \, L(\bfZ, \bfy; \bfalpha) + L(\bfW, \bfx; \bfbeta) \nonumber
\end{eqnarray}
where  $L(\bfZ, \bfy; \bfalpha) = \log(p(\bfZ| \bfy; \bfalpha))$,  $L(\bfW, \bfx; \bfbeta) = \log(p(\bfW| \bfx; \bfbeta))$, and where $\bfalpha$ and $\bfbeta$ symbolize all parameters in physician and patient feature distributions, respectively, and the second equal sign is because of the joint distribution can be factorized as in \ref{main}. Here we remark that since there is no unknown parameters in $p(\bfy| \bfx)$ and $p( \bfx)$, both of these two distributions do not play a role in the objective function. Because the objective function is separable, we can estimate $\bfalpha$ and  $\bfbeta$ by maximizing $L(\bfZ, \bfy; \bfalpha)$ and $L(\bfW, \bfx; \bfbeta)$ respectively. Next, we will derive the explicit form of the parameters.

\subsubsection*{Patient Feature Parameters}

By Equation \ref{para_main}, we compute the maximum likelihood estimates of the parameters and present the results in Table \ref{table:pat_para}. The table shows that the estimates from both classes are mostly close, which again reveals the challenge of our task. Recall that the parameters are used to determine the conditional distribution of patient features given patient labels, i.e., $p(\bfW| \bfx)$.

\begin{table}[!htb]
\centering
\captionsetup{justification=centering}
\caption{Maximum Likelihood Estimate of Patient Features \label{table:pat_para}}
\small
\begin{tabular}{l*{3}{c}r}
\hline
\hline
Parameter              & Negative class & Positive class  \\
\hline
Prob. of patient gender being male $\eta_g^0, \eta_g^1$            & 0.3812 &  0.2689 \\
Age decade distribution $\bfgamma_a^{0}, \bfgamma_a^{1}$   & See Figure \ref{age_dist} & \\
Region distribution $\bfgamma_e^{0}, \bfgamma_e^{1}$       &  [0.394, 0.223, 0.217, 0.166]  & \hspace{0.5cm}[0.316, 0.303, 0.194, 0.186]\\
Clinical code indicator parameters $\bfeta_d^{0}, \bfeta_d^{1}$     &  See Figure \ref{diag_flag_dist} & \\
Clinical code frequency parameters $\bflambda_d^{0}, \bflambda_d^{1}$     &   See Figure \ref{diag_freq_dist}  \\
\hline
\end{tabular}

\end{table}

In table \ref{table:pat_para}, the first line lists the parameter estimation for a patient gender being male in both negative and positive classes. From the result, we can see that the positive class patients have lower probability to have male gender (0.27 vs. 0.38). By the third line, the patient region distributions are close between two classes, which provides little information to differentiate the rare disease patients. This is a common phenomenon in a hard problem like this. We will need to utilize all information collectively to get better prediction.

\begin{figure}[!htb]
 \centerline
 {\includegraphics[width = 9 cm]{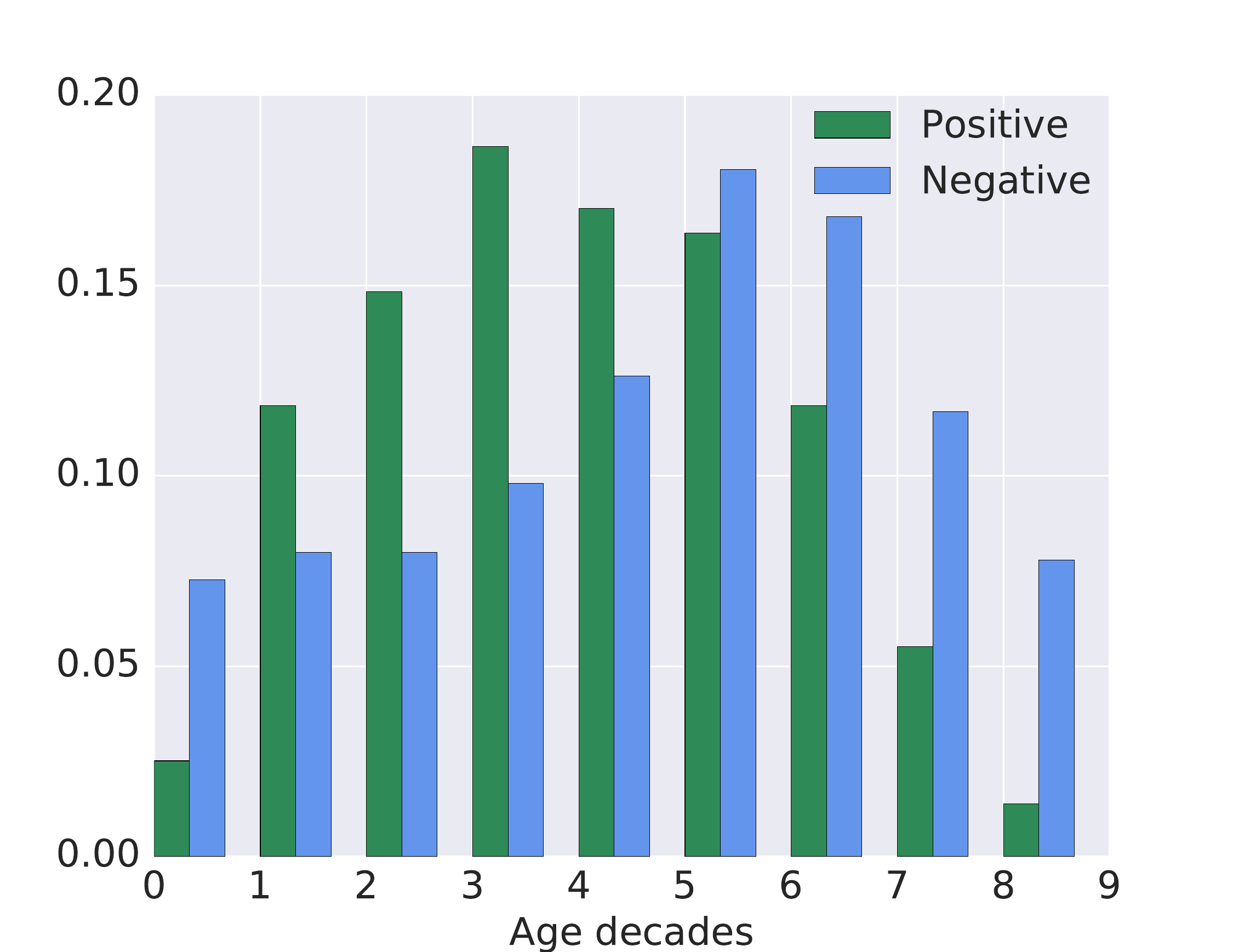}}
 \caption{Estimation results of $\bfgamma_a$,  patient age decade parameters}
 \label{age_dist}
\end{figure}

\begin{figure}[!htb]
 \centerline
 {\includegraphics[width = 9 cm]{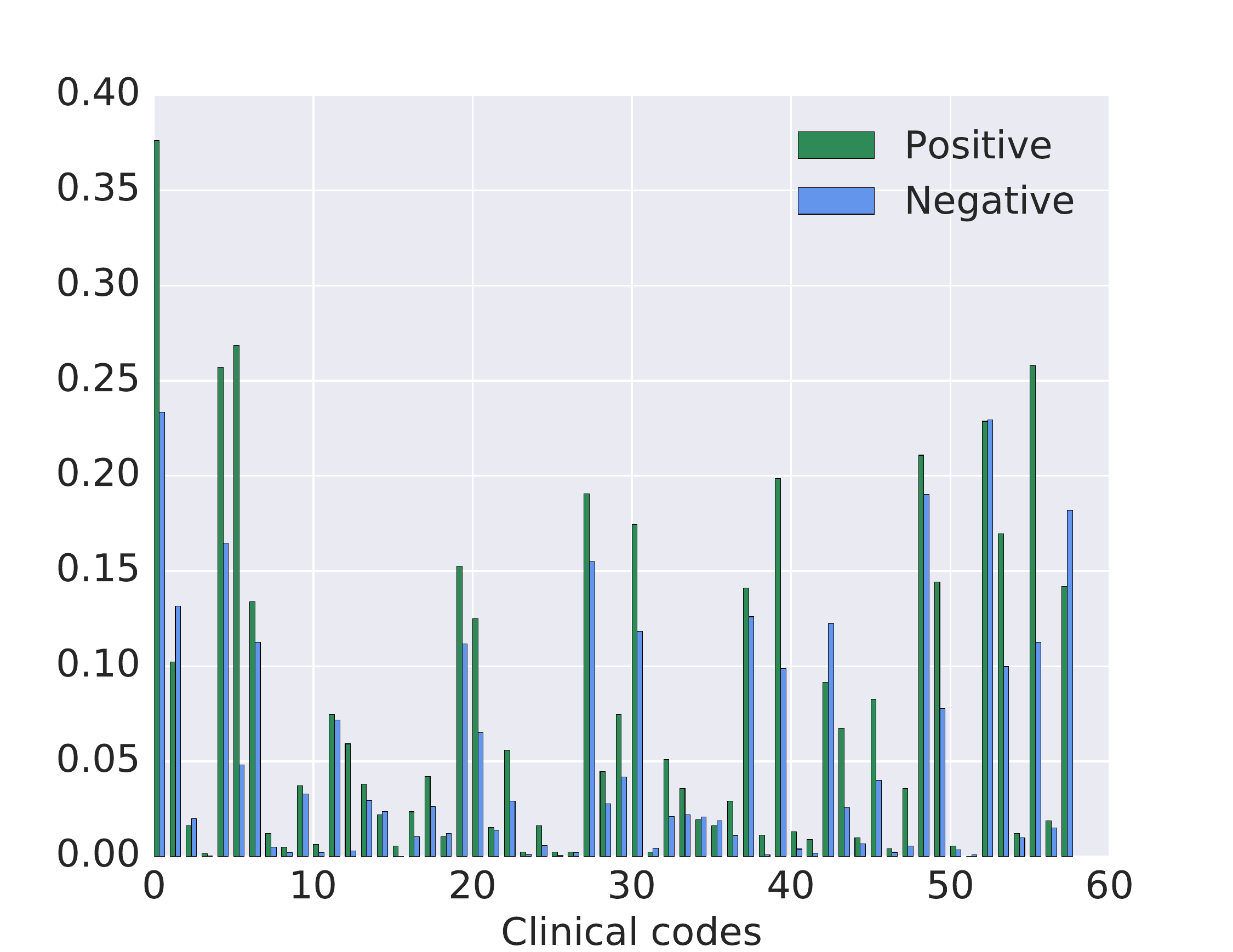}}
 \caption{Estimation results of $\bfeta_d$, patient clinical code indicator parameters. The x-axis index 0 to 57 represents the 58 types of clinical codes. }
 \label{diag_flag_dist}
 \end{figure}

Similarly we plot the probability mass function of patient age decades for both classes in Figure \ref{age_dist}. The age decade 0 denotes age from 0-9, 1 denotes 10-19, etc. The results from the training data set show that positive and negative classes have different distributions. If a patient is positive, then he or she is most likely to be in fifties than in other age decades, whereas if the patient is not positive, the age is most likely to be the thirties. At the same time, we observe that the ratio of positive patient among patients in thirties is  larger than this ratio of positive patients in fifties.

To demonstrate the results of patient clinical code indicator parameters $\bfeta$, we estimate the likelihood of predicting a positive label for both positive and negative class of patients. These estimates are obtained by computing the positive patient ratios in both classes. The results are plotted in Figure \ref{diag_flag_dist}, where we can see that for certain clinical codes, positive patients have a much higher probability to get a positive result than the negative patients. In particular, we list the estimated values of top five clinical codes with largest relative differences between classes in Table \ref{diag_table}. These top clinical codes represent clinical procedures, prescriptions and diagnoses.

\begin{table}[htb]
\centering
\captionsetup{justification=centering}
\caption{Clinical Codes Parameter Estimation - Top 5  \label{diag_table} }
\begin{tabular}{l*{2}{c}r}
\hline
\hline
Clinical codes              & Negative class $\eta_d^{0}$ & Positive class  $\eta_d^{1}$ \\
\hline
Chronic idiopathic urticaria         & 0.0031 &  0.0592 \\
Epinephrine                          & 0.0280 &  0.4184 \\
Personal history of Allergy          & 0.0054 &  0.0357 \\
Allergy/Anaphylaxis/Urticaria        & 0.0482 &  0.2685 \\
Laryngoscopy                         & 0.0152 &  0.0414 \\
\hline
\end{tabular}
\end{table}


\begin{figure}[htb]
 \centerline
 {\includegraphics[width = 9 cm]{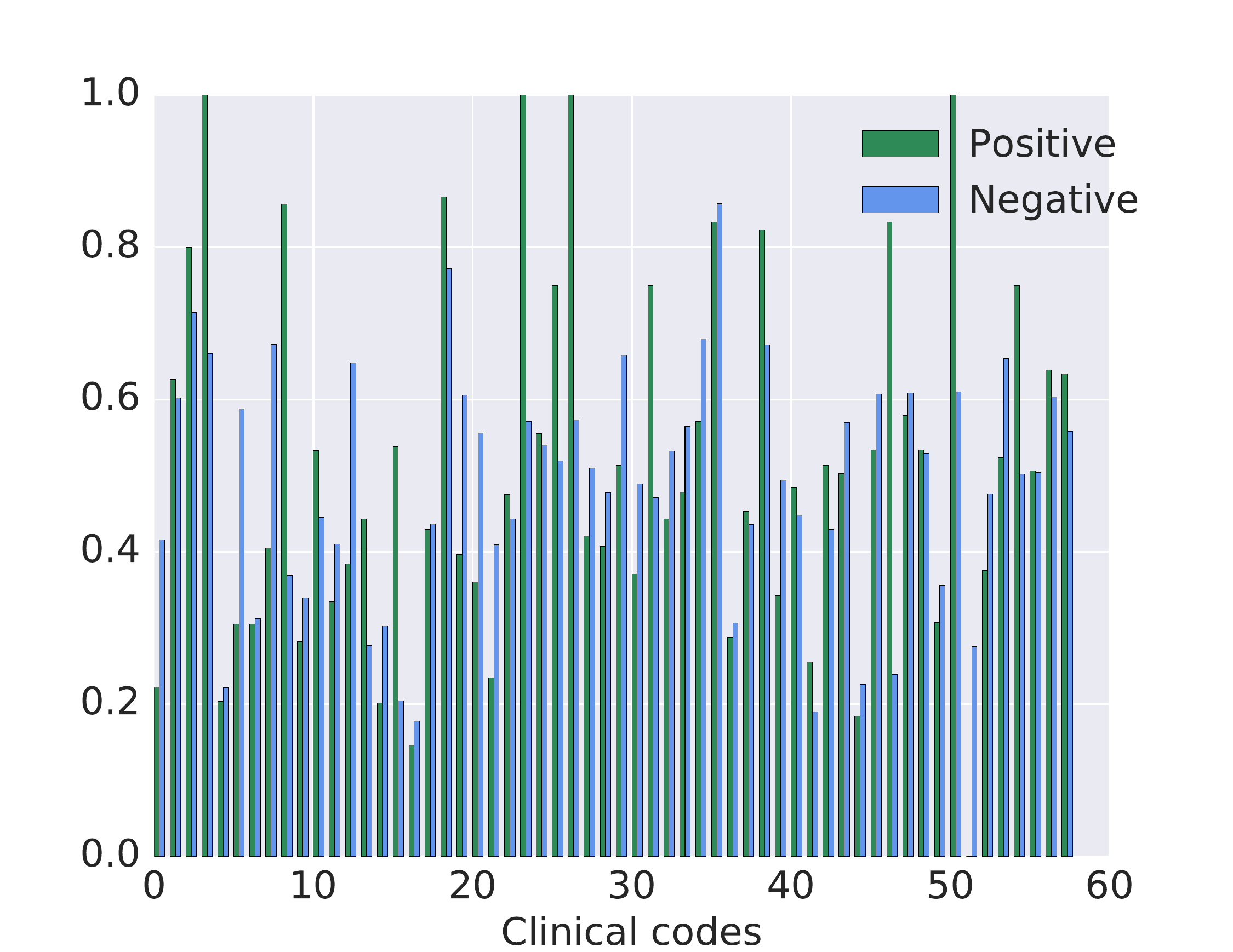}}
 \caption{Estimation result of $\bflambda_d$, clinical code frequency feature parameters, where the x-axis index from 0 to 57 represent the selected clinical codes. }
 \label{diag_freq_dist}
\end{figure}

The last line of parameters $\bflambda_d$ in Table \ref{table:pat_para} are two $58 \times 1$ column vectors. They are the parameters of the Poisson distribution which models the patient clinical code frequency features in each of the 58 clinical codes. Since the maximum likelihood estimate of Poisson distribution is the sample mean, we plot the parameter means for both positive and negative classes in  in Figure  \ref{diag_freq_dist}. We can find that the clinical code frequency features are similar for both positive and negative patients. This implies that patients with this rare disease condition look like all other patients in clinical code frequencies. Again these features may not have high predictive power by themselves.

\subsubsection*{Physician Feature Parameters}

We list physician parameter estimates in Table \ref{table:template}. These feature parameters include physician gender, patient count, specialty and number of office claims related. The first set of parameters $\eta_e^0$ and $\eta_e^0$ depicts the estimated probability of a physician gender to be male conditional on his positive or negative label. The second set of parameters $\lambda_c^0$ and $\lambda_c^1$ are for patient count distribution. Again, because of the Poisson assumption, these parameters can be computed using by the patient count mean for both of the classes. From the results, a physician who has positive label treats more patients (29) than a physician who hasn't (20).

\begin{table}[htb]
\centering
\captionsetup{justification=centering}
\caption{Physician Features Estimation Results \label{table:template}}

\renewcommand{\arraystretch}{1.2}
\begin{tabular}{r c}
\hline
\hline
Parameter              & Maximum likelihood estimate  \\
\hline
Prob. of physician gender being male $\eta_e^0$            & 0.8108   \\
$\eta_e^1$            & 0.7975   \\
Patient count mean $\lambda_c^0$   &  20.1514  \\
$\lambda_c^1$     &  29.0939  \\
Specialty distribution $\bfgamma_p^0$   &  See Figure \ref{spec_dist}  \\
$\bfgamma_p^1$ &     \\

Claims-features mean $\bfmu_d^0$     &  $[0.001 , 0.006 , 0.013 , -0.026]^\top$  \\
$\bfmu_d^1$     &  $[-0.007 , -0.042 , -0.097 , 0.191]^\top$  \\

Claims-features covariance $\bfSigma_D^0 $     & $\begin{bmatrix}
                            0.977 & 0.098 & 0.820 & 0.727 \\
                            0.098 & 1.108 & 0.246 & 0.136 \\
                            0.820 & 0.246 & 0.997 & 0.727 \\
                            0.727 & 0.136 & 0.727 & 0.897
                      \end{bmatrix}$  \\

$\bfSigma_D^1 $     & $\begin{bmatrix}
                            1.171 & 0.066 & 0.952 & 1.011 \\
                            0.066 & 0.216 & 0.085 & 0.055 \\
                            0.952 & 0.085 & 1.013 & 0.919 \\
                            1.011 & 0.055 & 0.919 & 1.704
                      \end{bmatrix}$ \\
\hline
\end{tabular}

\label{doc_para_res}

\end{table}

\begin{figure}[htb]
 \centerline
 {\includegraphics[width = 12 cm]{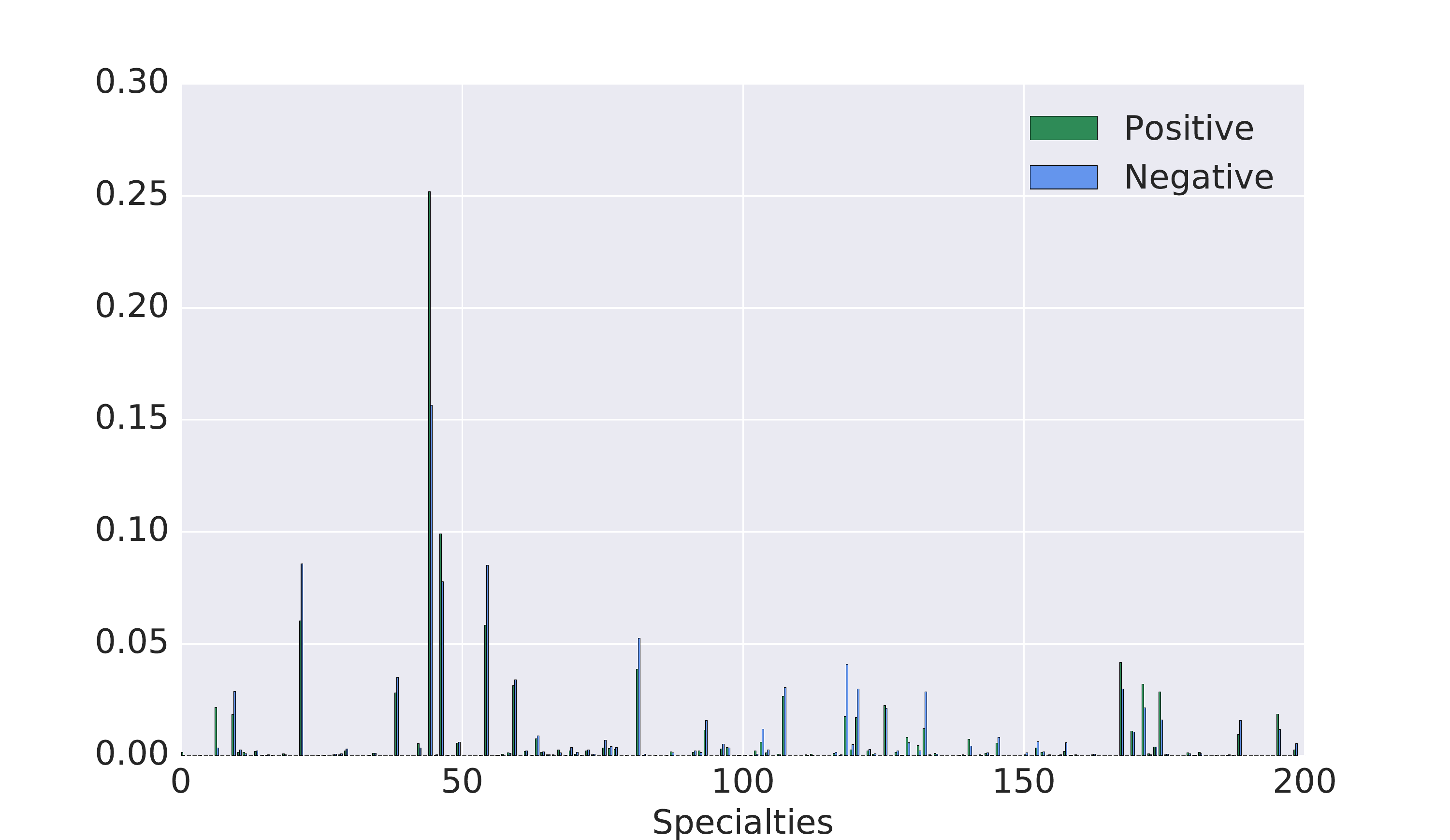}}
 \caption{Estimation result of $\bflambda_d$, physician specialty distribution, where the x-axis index 189 specialties. }
 \label{spec_dist}
\end{figure}

We plot all 189 specialties estimation results in Figure \ref{spec_dist}. Some of the specialty physicians are more likely to have this type of rare disease patient(s) than the others. Table \ref{table:doc_spec} lists the estimated value for the top 5 most common specialties. For example, among negative physicians 15.63\% of them are in specialty Diagnostic radiology, whereas among positive physicians, this ratio is 25.18\%.

\begin{table}[htb]
\centering
\captionsetup{justification=centering}
\caption{Physician Specialty Estimation - Top 5 Most Common\label{table:doc_spec}}

\renewcommand{\arraystretch}{1.2}
\begin{tabular}{r c c}
\hline
\hline
Specialty              & Negative class & Positive class  \\
\hline
Diagnostic radiology            & 15.63\% &  25.18\% \\
Emergency medicine            & 7.78\% &  9.91\% \\
Cardiovascular disease            & 8.57\% &  6.02\% \\
Family medicine            & 8.51\% &  5.83\% \\
Anatomic/clinical pathology            & 2.97\% &  4.18\% \\
\hline
\end{tabular}

\end{table}


In the physician feature distribution section, we assume the  claims related features such as maximum, minimum, sum and average number of claims follow a multi-variate Gaussian distribution. So the sample mean and the sample covariances are the maximum likelihood estimates of the  $\bfmu_d$ and $\bfSigma_D $. We show the results in the last four lines of Table \ref{table:template}. There is no direct interpretation for these estimation results. But collectively with all other information they can contribute to the prediction improvement.

\section*{Empirical Results}

To validate the proposed model, we will provide two experiments that show the performance of our method and its comparisons to the random forest method \cite{ho1998random}. In the experiments, we use positive predictive value(PPV) vs. sensitivity analysis \cite{friedman2001elements}, F1 score and the Matthews correlation coefficient for performance evaluation.

Since we form this rare disease targeting as a binary classification problem, the detector performance can be measured by the true positive(TP), false positive(FP), true negative(TN) and false negative(FN). Then the PPV (also referred as  precision) is defined as PPV = TP / (TP + FN); and sensitivity (also called recall) is defined as sensitivity = TP / (TP + FP). Namely, PPV is the number of correct positive results divided by the number of all positive results, and sensitivity is the number of correct positive results divided by the number of the predicted positive results. When comparing two detectors, with same sensitivity, the better detector should have a larger PPV value.  Similary, with identical PPV, the detector with larger sensitivity performs better. Obviously, both the PPV and sensitivity are between zero and one.

Another performance metric of binary classification's accuracy is the F1 score, which summarizes the PPV and sensitivity to a single real number. The F1 score is defined as two times the harmonic mean of PPV and sensitivity, given by $ \dfrac{2 \cdot \mathrm{PPV} \cdot \mathrm{sensitivity}}{\mathrm{PPV} + \mathrm{sensitivity}} $. F1 score have value between 0 and 1, where an F1 score reaches its best value at 1 and worst at 0.

The third performance metric we use is the Matthews correlation coefficient(MCC)\cite{matthews1975comparison}. It is  generally regarded as a balanced performance measure for binary classification which can be used even if the classes are of very different sizes. A coefficient equal to 1 means perfect prediction, 0 denotes random guess and −1 indicates total disagreement between prediction and observation. Noting that the data set used for validation is highly imbalanced, we use MCC as the third performance metric for its consistency in data balance.

We select the random forest model as our benchmark method for results comparison. The random forest classifier has demonstrated its capability and robustness in a variety of classification problems \cite{liaw2002classification,diaz2006gene,svetnik2003random}. For this benchmark, we specify the random forest with 200 decision trees. The random forest model can use the same physician level features. But unlike the proposed model, it cannot incorporate patient and physician dependency directly. For each physician, we average all patients records that link to this physician and create similar patient features at physician level. The benchmark random forest is implemented through the Python package scikit-learn \cite{scikit-learn}.

\subsubsection*{One-fold validation}

In the first experiment, we split randomly all the 68,898 physicians into one testing and one training data sets. The training set has 6,000  ( $10\%$ ) physicians, where 713 physicians are positive and 5,287 are negative. The testing set contains the rest 62,898 physicians including 55,265 positive and 7,633 negative physicians respectively. To avoid information leakage from the patient label in the training data set, any patient connected with any physician in testing data set shall be regarded as a testing patient. As a result, according to the train-test split in the physicians, 161,681 patients are grouped into the training set, and 86,152 patients are grouped into the testing set.  There are 736 positive and 160,945 negative patients in the training set, and there are 497 positive and 85,655 negative patients.

In Table \ref{table:TTS result}, we show the experiment result of the proposed result and the benchmark method, where the subscript `pm' and `bm' correspond to `proposed method' and `benchmark method' respectively.  We can see that the proposed method shows a much higher PPV than that of benchmark method. In particular, although the benchmark method hardly work (PPV less than 0.05) when the sensitivity is greater than 0.35, the proposed method yields an acceptable PPV (greater than 0.2). Similarly, the F1 score by the proposed method consistently higher than that of the benchmark method, especially when the sensitivity is greater than 0.35. In comparison with the best MCC of the benchmark method 0.3699, the best MCC of the proposed method is  0.4207, which implies a $13.7\%$ performance increase.

\begin{table}[htb]
\centering
\captionsetup{justification=centering}
\caption{Comparison results between proposed and benchmark methods in one-fold validation \label{table:TTS result}}

\begin{tabular}{l*{7}{c}r}
\hline
\hline
Sensitivity  & $\textrm{PPV}_{pm}$ & $\textrm{PPV}_{bm}$ & $\textrm{MCC}_{pm}$ & $\textrm{MCC}_{pm}$ & $\textrm{F1 score}_{\,pm}$  & $\textrm{F1 score}_{\,bm}$\\
\hline
0.2            & 0.8036  & 0.7826 &  0.2403  & 0.2354  & 0.3206  & 0.3190 \\
0.25           & 0.7013  & 0.5946 &  0.2860  & 0.2549  & 0.3681  & 0.3523 \\
0.3            & 0.6143  & 0.4825 &  0.3168  & 0.2720  & 0.4026  & 0.3699 \\
0.35           & 0.4937  & 0.0589 &  0.3219  & 0.1021  & 0.4207  & 0.1008 \\
0.4            & 0.2917  & 0.0449 &  0.2686  & 0.1010  & 0.3379  & 0.0807 \\
0.45           & 0.2019  & 0.0365 &  0.2429  & 0.1006  & 0.2788  & 0.0674 \\

\hline
\end{tabular}

\label{pat_para_res}

\end{table}

\begin{figure}[htb]
 \centerline
 {\includegraphics[width = 12 cm]{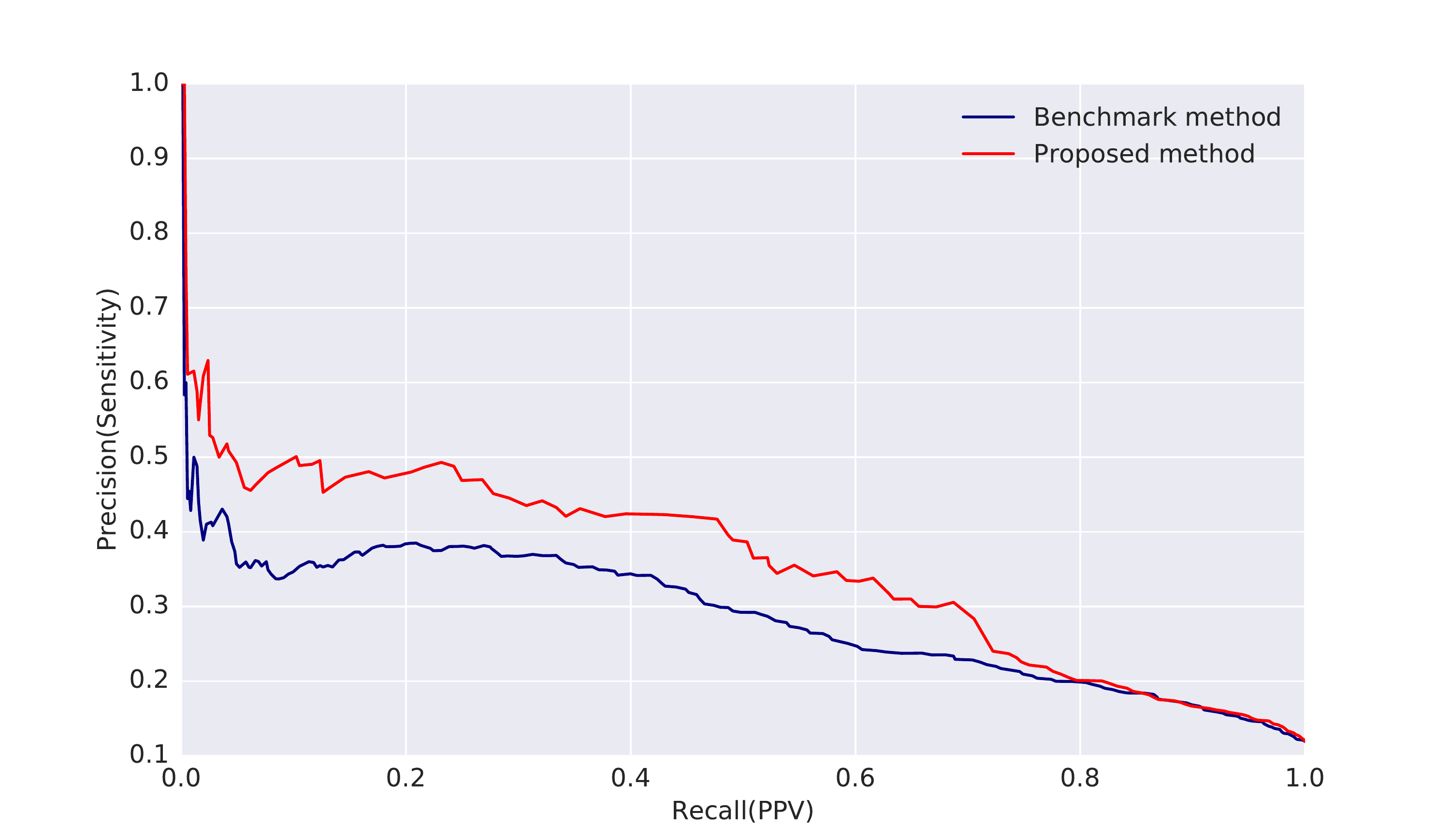}}
 \caption{Sensitivity with respective to PPV curves for proposed and benchmark methods in one-fold validation }
 \label{prcurve}
\end{figure}

To compare the two methods in term of PPV values,  in Figure \ref{prcurve}, we plot the sensitivity-PPV curves of both methods, where each dot on the curve represents a sensitivity PPV pair. We can see almost for every value of PPV, the proposed method has a higher sensitivity than the benchmark method. The area under curve serves as a single variable summary of the PPV performance, which is 0.3553 for the proposed method, and 0.2857 for the benchmark method, suggesting that the overall relative performance increase is $24.31\%$.

\subsubsection*{Ten-fold cross validation}

In the second experiment, we carry out the ten-fold cross validation with both methods to demonstrate the robustness of the proposed method. Specifically,  the physicians are randomly partitioned into 10 equal size groups. Then the process of the first experiment is repeated for 10 times, where in each time one single group of the physicians is retained as the testing set, while all the rest 9 groups are treated as training data set. Note that there is no overlap of any two testing data set. Consequently, 10 set of performance measure results are obtained. Table \ref{table:TTS result} shows the average PPV and MCC of both methods with respect to various sensitivity values. We can see that the proposed method outperforms the benchmark in term of both average PPV and average MCC.

\begin{table}[htb]
\centering
\captionsetup{justification=centering}
\caption{Comparison results between proposed and benchmark methods in ten-fold validation \label{table:TTS result}}

\begin{tabular}{l*{5}{c}r}
\hline
\hline
$\,$ & ${\textrm{Average PPV}}$ & $\,$ & $\textrm{Average MCC}$ & $\,$ \\
\hline
Sensitivity  & Proposed method &  Benchmark method & Proposed method & Benchmark method \\
\hline
0.2            & 0.8046  & 0.6469 & 0.2354 & 0.1943   \\
0.25           & 0.6839  & 0.4664 & 0.2781 & 0.2132   \\
0.3            & 0.5732  & 0.3342 & 0.3007 & 0.2158   \\
0.35           & 0.4261  & 0.0754 & 0.2893 & 0.1019   \\
0.4            & 0.2332  & 0.0690 & 0.2282 & 0.1153   \\
0.45           & 0.1295  & 0.0441 & 0.1821 & 0.0991   \\

\hline
\end{tabular}

\label{pat_para_res}

\end{table}

\begin{figure}[htb]
 \centerline
 {\includegraphics[width = 12 cm]{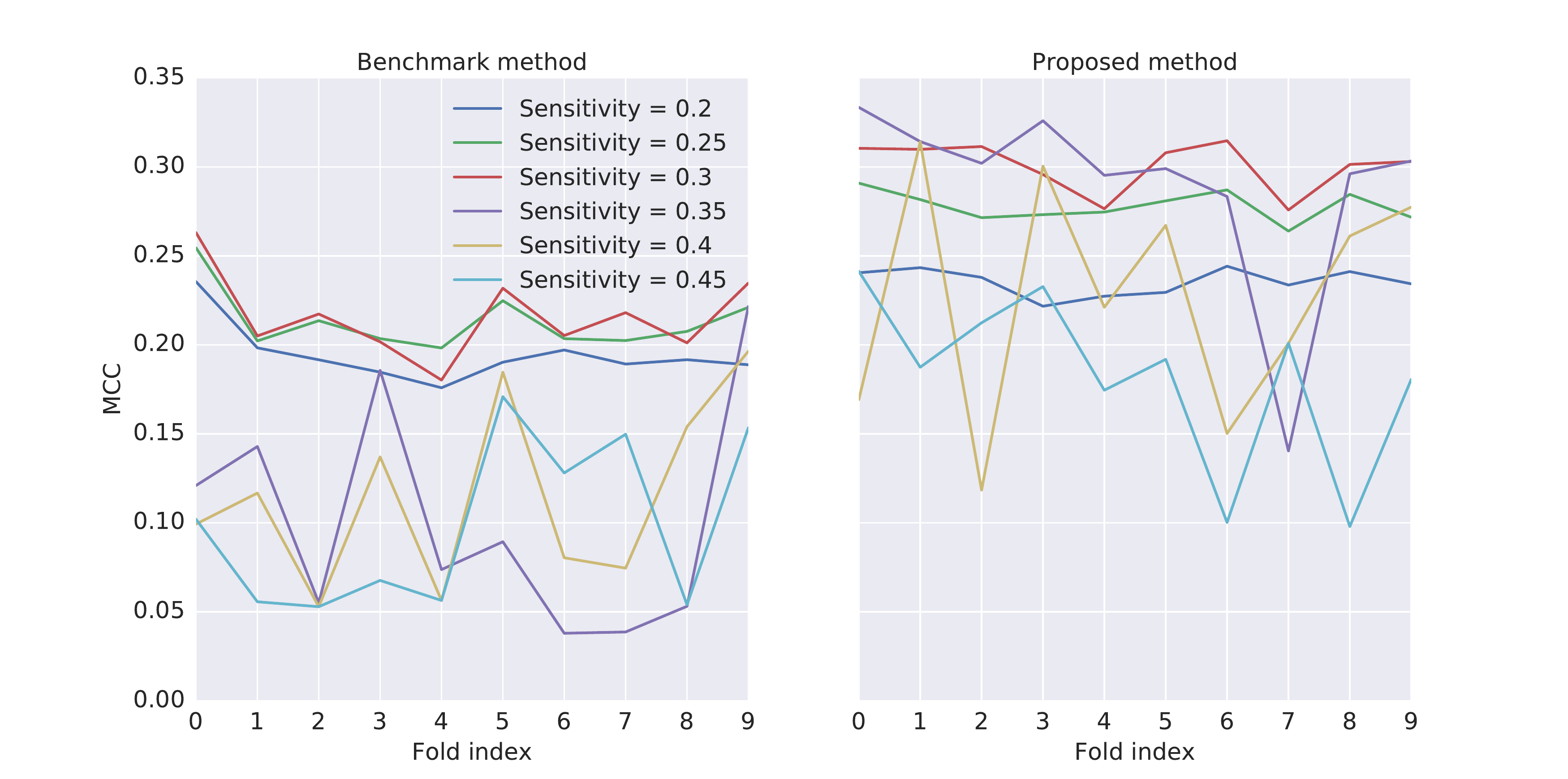}}
 \caption{Ten-fold MCC results of the benchmark and proposed methods}
 \label{tenfold_MCC}
\end{figure}

For more detailed comparison results, in Figure \ref{tenfold_MCC} we plot the ten fold results for both methods, where the the x-axis denotes the fold index from 0 to 9, the y-axis represents the MCC values. Comparing the left and right sub-figures, we can see that with identical color, i.e., with same sensitivity level, the curve on the right sub-figure is higher than that on the left figure, which shows that in all ten folds, the proposed method has a higher MCC.

\section*{Conclusion}

The goal of this article is to enhance rare disease physician targeting precision by exploiting extra structural relationship among physicians and patients. It is often a hard problem to identify rare disease treating physicians out of a large physician population. The difficulties come from many aspects such as extreme imbalance in classes and rare disease patients looking alike to common condition patients. We propose a graphic representation and probability model to join physician and patient features together with their network relationship. Through the graphical structure, researchers can visualize the connectivity among physicians and patients. The graphic representation provides clear interpretability of data entities and correlation. The proposed model also has flexibility to specify additional dependencies, add features or extend to more complicated network structure. In the empirical example, we use factor graph to predict physician rare disease flag. We compare the results to random forest benchmark and find much improved targeting accuracy. Especially at high sensitivity level, the proposed method show significant improvement over benchmark. In practice, this means when a smaller target is needed under tight marketing budget, the proposed method can yield superior results by identifying more real targets.

The literature shows that the graphic model methods such as factor graph or Markov random fields have the ability to specify and utilize these relationship to improve performance. In pharmaceutical marketing, there exists complex relationship among various stakeholders. But to our knowledge, there is limited effort to take advantage of such information. This article provide extra data point to demonstrate the usefulness of utilizing physician and patient structural link.

\subsubsection*{Future research}
The presented case has certain limitations. First of all, we use binary classifier to predict rare disease physician identity. The outcome only indicates whether or not a physician having rare disease patients. The future research can extend to multi-class or continuous response such that it can predict how many rare disease patients for a predicted physician.

In this study we use independent assumption among some of the feature to simplify model structure. We hope other studies can consider adding covariance to account for such feature correlations. Furthermore, we only include basic physician features such as specialty and location, etc. in our demonstration. But there are extensive historical treatment data available for each physician. It is possible to enhance the physician similarity link by incorporating those information in the graph nodes. A stronger physician similarity structure can potentially increase predictive power. It also calls for extra research on how to extra useful features from high dimensional physician history data.

Although in our study the proposed model improves rare disease targeting accuracy to certain degree, the precision still leaves a lot to be desired. We hope to see more innovative methods to tackle this hard problem in the future research.

\newpage

\bibliographystyle{IEEEbib}
\bibliography{refs}

\end{document}